\pdfoutput=1

\documentclass[11pt]{article}

\usepackage[final]{acl}

\usepackage{times}
\usepackage{latexsym}
\usepackage{multirow}
\usepackage{booktabs}
\usepackage{amsmath}
\usepackage{amssymb}
\usepackage{adjustbox}
\usepackage{amsfonts}

\usepackage{cleveref}
\usepackage{graphicx}
\usepackage{epstopdf}
\usepackage{hyperref}
\usepackage{colortbl}
\usepackage{float}
\usepackage{pifont}
\usepackage{xcolor}
\usepackage{mdframed}

\usepackage{tabularx}

\definecolor{my_green}{RGB}{51,102,0}
\definecolor{my_red}{RGB}{204, 0, 0}
\makeatletter
\let\scshape\relax 
\DeclareRobustCommand\scshape{%
  \not@math@alphabet\scshape\relax
  \ifnum\pdf@strcmp{\f@family}{\familydefault}=\z@
    \fontfamily{qbk}%
  \fi
  \fontshape\scdefault\selectfont}
\makeatother

\usepackage[T1]{fontenc}
\usepackage[utf8]{inputenc}
\usepackage{microtype}
\usepackage{inconsolata}
\usepackage{graphicx}

\title{Reinforced Information Retrieval}

\author{
    Chaofan Li$^{1,2}$\thanks{~ The two authors contribute equally.}, \ Zheng Liu$^{1,4*}$, \ Jianlyu Chen$^{1,3}$, 
    \ \textbf{Defu Lian}$^{3}$, \ \textbf{Yingxia Shao}$^{2}$  \\
    $^1$  BAAI,  \ \ \ \ $^2$ BUPT, \ \ \ \ $^3$  USTC, \ \ \ \ $^4$ HKPU  \\
    \texttt{zhengliu1026@gmail.com}  \quad \texttt{\{cli,yxshao\}@bupt.edu.cn}
}

\begin{document}
\maketitle
\begin{abstract}


While retrieval techniques are widely used in practice, they still face significant challenges in cross-domain scenarios. Recently, generation-augmented methods have emerged as a promising solution to this problem. These methods enhance raw queries by incorporating additional information from an LLM-based generator, facilitating more direct retrieval of relevant documents.  However, existing methods struggle with highly specialized situations that require extensive domain expertise. To address this problem, we present \textbf{Reinforced-IR}, a novel approach that jointly adapts a pre-trained retriever and generator for precise cross-domain retrieval. A key innovation of Reinforced-IR is its \textbf{Self-Boosting} framework, which enables retriever and generator to learn from each other's feedback. Specifically, the generator is reinforced to generate query augmentations that enhance the retriever's performance, while the retriever is trained to better discriminate the relevant documents identified by the generator. This iterative process allows the end-to-end retrieval performance to be progressively optimized using an unlabeled corpus from the target domain. In our experiment, Reinforced-IR outperforms existing domain adaptation methods by a large margin, leading to substantial improvements in retrieval quality across a wide range of application scenarios.   

\end{abstract} 

\section{Introduction} 
With the rapid advancement of large language models (LLMs), AI copilots have become deeply integrated into a wide variety of activities, such as addressing knowledge-intensive problems, analyzing professional documents, developing computer programs, and providing personal assistance \cite{achiam2023gpt,team2023gemini,anthropic2024claude}. To produce reliable and trustworthy results in these tasks, it is essential to incorporate useful knowledge from external databases, a process known as retrieval-augmented generation of LLMs, i.e., RAG \cite{lewis2020retrieval}. Because of the advantages in broad applicability and simplicity, dense retrieval emerges as a popular form of retriever in such applications. It employs an embedder to map the data into a vector space, enabling the retrieval of relevant information based on vector similarity \cite{zhao2024dense}. Recently, numerous open-source models and API services have been made publicly available \cite{izacard2021unsupervised,xiao2024c,neelakantan2022text}, which significantly facilitate the utilization of corresponding techniques.   

Given the diverse range of applications, it's important to adapt general retrievers to new working scenarios beyond their original training domains. To this end, a variety of domain adaptation methods have been proposed in recent years. A notable breakthrough was made by the development of \textit{HyDE}-style methods (hypothetical document embedding) \cite{gao2022precise,wang2023query2doc}, or more broadly, the \textit{GAR} techniques (generation-augmented retrieval) \cite{mao2020generation}. These methods leverage LLMs, like ChatGPT, to enrich the query with extra information, thus enabling relevant documents to be identified in a straightforward way. \textit{However, the existing methods mainly rely on LLMs trained on general domains, which may lack necessary knowledge needed by a highly specialized domain, such as medical or legal retrieval.} Besides, the heavy reliance on proprietary LLMs often results in prohibitively high costs, which limits their applicability in many situations.  

\begin{figure*}[tb]
    \centering
    \includegraphics[width=0.95\textwidth]{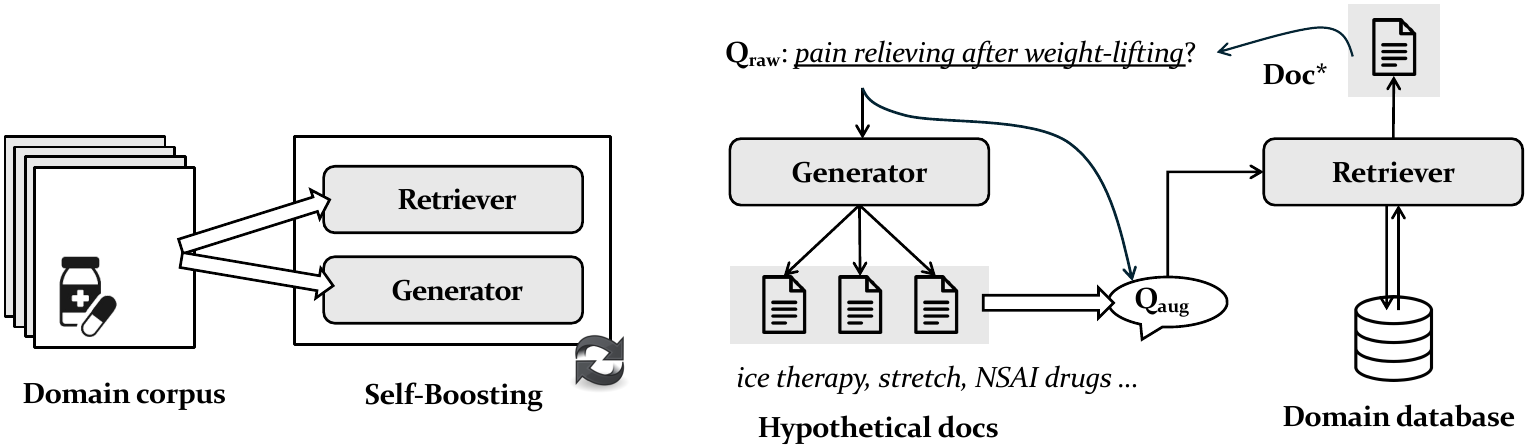}
    \vspace{-0.3cm}
    \caption{\textbf{Reinforced-IR} jointly adapts retriever and generator with an unlabeled domain corpus via self-boosting. The well-adapted generator augments raw query with hypothetical docs, which enables relevant docs to be retrieved.} 
    \vspace{-0.5cm}
    \label{fig:frame}
\end{figure*}

To address these challenges, we propose a novel domain adaptation framework called \textbf{Reinforced-IR}, \textit{which jointly adapts the {retriever} and {LLM-based generator} using a {unlabeled corpus}}. Our method is distinguished for its design of \textbf{self-boosting} algorithm. It starts with a list of pseudo questions generated from the target domain's unlabeled corpus. On one hand, the LLM-based generator is reinforced to perform high-quality query augmentation using the retriever's feedback, such that relevant documents can be optimally retrieved for downstream tasks. This step is referred as the \textit{Reinforcement Learning of generator with Retriever's Feedback} (\textbf{RLRF}). On the other hand, the retriever is reinforced to discriminate the relevant documents preferred by the LLM-based generator. This step is called the \textit{Reinforcement Learning of retriever with Generator's Feedback} (\textbf{RLGF}). With the alternating execution of these two operations, the end-to-end retrieval performance can be progressively enhanced for the target domain.



We perform a comprehensive evaluation based on a variety of domain-specific datasets from BEIR \cite{thakur2021beir} and AIR-Bench \cite{chen2024air}. We also include various retrievers and LLMs in our evaluation. According to the experiment result, Reinforced-IR substantially enhances the cross-domain performance of pre-trained retrievers and demonstrates notable advantages over the existing domain adaptation baselines. Additionally, the performance gains are especially pronounced on low-resource datasets that differ substantially from the original domains of the retrievers and LLMs, which further highlights the effectiveness of our approach for domain adaptation. Our model and source code will be shared with the public to advance the future research in this field.   

In summary, the contributions of this paper are presented as follows:
\begin{itemize}
    \item We introduce Reinforced-IR, a novel framework for cross-domain retrieval. To the best of our knowledge, this is the first work that jointly adapts retriever and generator to optimize end-to-end retrieval performance.      
    \item We design the \textbf{RLRF} and \textbf{RLGF} algorithms, enabling retriever and generator to mutually enhance each other’s performance based on an unlabeled corpus from the target domain.  
    \item We conduct comprehensive experimental studies, which verify our significant advantage over existing cross-domain retrieval methods. 
\end{itemize}


\begin{figure*}[tb]
    \centering
    \includegraphics[width=0.90\textwidth]{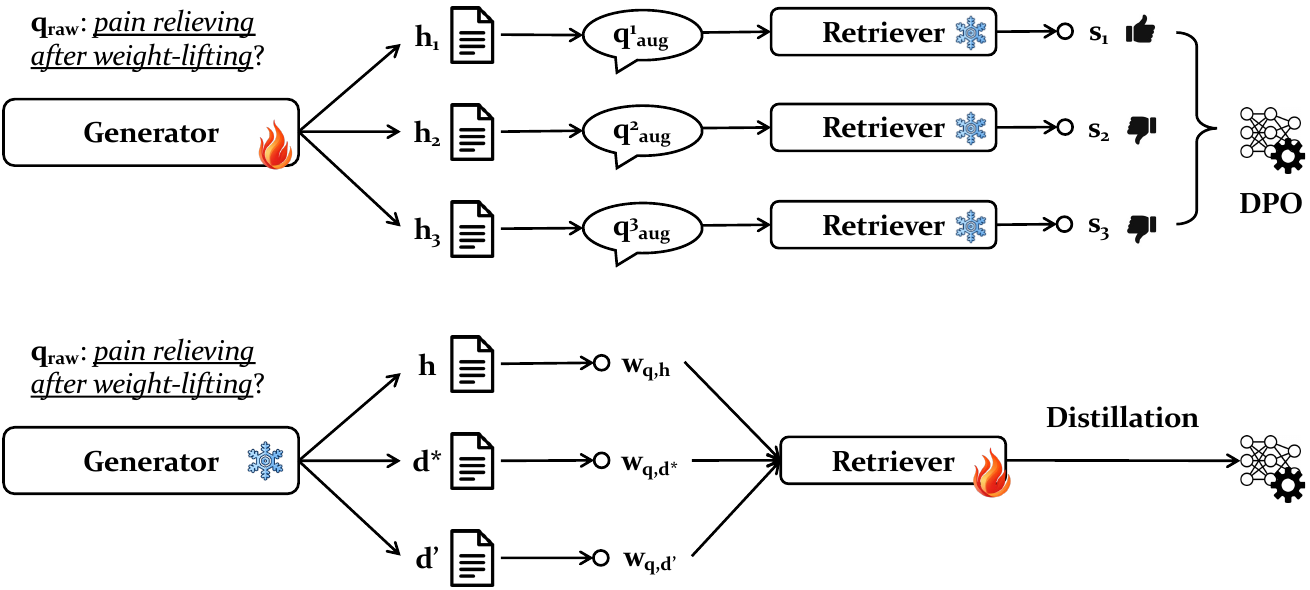}
    \vspace{-0.3cm}
    \caption{\textbf{Self-Boosting workflow}. 1) \textbf{RLRF}: the generator is reinforced to produce the retriever's preferred query augmentation (marked by thumb-up) through DPO. 2) \textbf{RLGF}: the retriever is reinforced to discriminate the generator's preferred documents (measured by preference score $w_{-}$) in the form of knowledge distillation.} 
    \vspace{-0.3cm}
    \label{fig:method}
\end{figure*}

\section{Method}
In this section, we will first introduce the workflow of generation-augmented retrieval and formulate the problem. Then, we will elaborate the self-boosting algorithm, which optimizes the end-to-end retrieval performance using unlabeled data. 

\subsection{Generation-Augmented Retrieval} 
As a popular IR paradigm, dense retrieval identifies a query's relevant documents based on embedding similarity. Given an embedding model $enc(\cdot)$, the query $q$ and document $d$ are transformed into latent vectors: $\boldsymbol{v}_q \leftarrow enc(q)$, $\boldsymbol{v}_d \leftarrow enc(d)$. On top of such results, the relevance score is calculated as the following inner product: $\sigma_{q,d} \leftarrow \boldsymbol{v}_q^T \boldsymbol{v}_d$. It is expected that the most relevant document ($d^*$) can produce the highest relevant score compared to the rest of documents, i.e., $d^*: \max \{ \boldsymbol{v}_q^T \boldsymbol{v}_d \}_{d \in D}$. 

When applied to a new scenario, the model needs to handle different relevance patterns between query and document from its original domain. To bridge this gap, the query is augmented with extra information (Figure \ref{fig:frame}), like hypothetical docs in HyDE \cite{gao2022precise}. Despite possible incomplete or inaccurate details, the generation-augmented retrieval (GAR) facilitates query and relevant docs to be matched in a more straightforward way. Nowadays, the query augmentation is often performed by a LLM-based generator $gen(\cdot)$, which are directly prompted to generate a list of hypothetical docs ($H_q$) for the query: 
\begin{equation}
    H_q: \{ h_i \leftarrow gen(q, ~prompt) \}_{i=1,...,L}.
\end{equation}
Here, $h_i$ is one of the sampled generation results. The system prompt is defined w.r.t. each concrete scenario, e.g., ``\textit{\{Query\} (symptoms of some disease). Generate the treatment for the described disease}'' for a medical retrieval problem. Following the proposed workflow in HyDE, the augmented query embedding ($\boldsymbol{v}'_{q}$) is calculated as the linear combination of raw query embedding ($\boldsymbol{v}_q$) and each of the hypothetical document embeddings: 
\begin{equation}\label{eq:2}
    \boldsymbol{v}'_{q} \leftarrow \alpha_0*\boldsymbol{v}_q + \sum\nolimits_{1...L} \alpha_i * \boldsymbol{v}_{h_i}, 
\end{equation} 
where $\boldsymbol{v}_{h_i} \leftarrow enc(h_i)$, $\alpha_i>0$, and $\sum_{0...L}\alpha_i=1$. Ultimately, the augmented query embedding $\boldsymbol{v}'_{q}$ is used for the retrieval of relevant documents. 

With the above definition, our problem is formulated as the joint optimization of embedding and generation model: $enc(\cdot)$, $gen(\cdot)$, such that the relevant documents in the target domain can be identified using the augmented query embedding, i.e., $d^*: \max \{ \boldsymbol{v}'^T_q \boldsymbol{v}_d \}_{d \in D}$. 


\subsection{Self-Boosting} 
The optimization process begins with a unlabeled corpus ($D$) from the target domain. Following established practices \cite{ma2020zero,thakur2021beir}, we prompt the LLM to generate a set of synthetic queries: $Q \leftarrow \{q: QGen(d^*) \}_{D^*}$ for sampled documents $D^*$. The resulting pairs $\{(q, d^*)\}_Q$ serve the training source for domain adaptation. Building on this foundation, we introduce the \textbf{Self-Boosting} algorithm, which consists of two dual steps: {generator optimization} by \textit{reinforcement learning from retriever's feedback} (\textbf{RLRF}), and {retriever optimization} by \textit{reinforcement learning from generator's feedback} (\textbf{RLGF}). The two steps are iteratively performed, enabling progressive improvement in end-to-end retrieval performance. 

\subsubsection{Generator optimization by RLRF} 
The generator is prompted to produce a group of candidate hypothetical documents for each training query $q$: $H_q \leftarrow \{h_i \leftarrow gen(q, prompt)\}_{i=1,...,K}$, where $K$ is the predefined sample size. Given this sampling result, the generator is reinforced to produce the best candidate which optimizes the retriever's performance. Particularly, we simplify the calculation of augmented query embedding in Eq. \ref{eq:2} as the case with one single hypothetical document: $\boldsymbol{v}'_q \leftarrow \alpha *\boldsymbol{v}_q + (1-\alpha)*\boldsymbol{v}_{h_i}$, where we compute the retriever's preference score as: 
\begin{equation}\label{eq:4}
    s_{q,h_i} \leftarrow \boldsymbol{v}'^T_q \boldsymbol{v}_{d^*} . 
\end{equation}  
By applying the above computation to every augmented query, we get $S_q \leftarrow \{s_{q,h_i}\}_{H_q}$ as the retriever's feedback to the whole hypothetical documents. We further conduct direct preference optimization (DPO) to reinforce the generation of the retriever's preferred query augmentation \cite{rafailov2024direct}. For the simplicity of training, we only consider the hypothetical documents of the highest and lowest scores, and leverage them as the wining and losing candidates: $h_w$, $h_l$. To screen out low-quality samples, we introduce the following filtering rules to the candidate documents: 
\begin{equation}
    1. ~ s_{q,h_w} > s_{q}, \ \, 2. ~ s_{q,h_w} > \gamma * s_{q,h_l},
\end{equation}
where $\gamma$ is a scaling factor: $\gamma>1$, $s_{q}$ indicates the preference score without using hypothetical document: $s_{q} = \boldsymbol{v}_q^T \boldsymbol{v}_{d^*}$. The first rule regularizes that the winning candidate must positively contribute to the retrieval result, while the second one guarantees the significance of winning candidate's contribution. Finally, we apply the following loss for DPO: 
\begin{equation}
    \mathcal{L}^{dpo} = - \log \sigma\big(\beta\log\frac{\pi(h_w|q)}{\pi'(h_w|q)} - \beta\log\frac{\pi(h_l|q)}{\pi'(h_l|q)} \big), 
\end{equation} 
where $\pi$ and $\pi'$ are the conditional likelihood from the adapted generator and the original generator respectively, and $\sigma(\cdot)$ is the sigmoid function.  

\subsubsection{Retriever Optimization by RLGF}
The retriever needs to make effective use of the augmented query. To this end, we maximize the relevance score between $\boldsymbol{v}'_q$ and $\boldsymbol{v}_{d}$: $\boldsymbol{v}'^T_q \boldsymbol{v}_{d}$. As $\boldsymbol{v}'_q$ is a linear combination of multiple embeddings (Eq. \ref{eq:2}), the following decomposition is made: $\alpha_0 * \boldsymbol{v}_q^T \boldsymbol{v}_{d} + \sum_{L} \alpha_i * \boldsymbol{v}_{h_i}^T \boldsymbol{v}_{d}$. However, optimizing this objective involves two different capabilities from the retriever: 1) \textit{query-to-doc} matching as required by $\boldsymbol{v}_q^T \boldsymbol{v}_{d}$, 2) \textit{doc-to-doc} matching as needed by $\boldsymbol{v}^T_{h_i} \boldsymbol{v}_{d}$. Thus, the direct optimization process is challenging, as it must realize two distinct goals simultaneously. To address this problem, we propose the \textit{proximity objective} ($\rho_{q,d}$) as an alternative: 
\begin{equation}\label{eq:6}
    \rho_{q,d} = \alpha_0 * \boldsymbol{v}_q^T \boldsymbol{v}_{d} + \sum\nolimits_{L} \alpha_i * \boldsymbol{v}_q^T \boldsymbol{v}_{h_i}.
\end{equation} 
The proximity objective maximizes the embedding similarity between the query and hypothetical documents, i.e., $\boldsymbol{v}_q^T \boldsymbol{v}_{h_i}$. Consequently, the retriever focuses solely on the \textit{query-to-doc} matching capability, which makes it easier to optimize. In addition, the above objective leverages $\boldsymbol{v}_q$ as an anchor where both $\boldsymbol{v}_{d}$ and $\boldsymbol{v}_{h_i}$ are moved close to it. Therefore, the similarity between $\boldsymbol{v}_{d}$ and $\boldsymbol{v}_{h_i}$ can also be improved from its optimization. Based on the above definition, we initially formulate the contrastive loss for retriever's training: 
\begin{equation}
    \mathcal{L}^{ctr} = -  \sum\nolimits_{D^*_q} \log \frac{ \exp( \boldsymbol{v}_q^T \boldsymbol{v}_{d} )}{\sum\nolimits_{D'_q} \exp( \boldsymbol{v}_q^T \boldsymbol{v}_{d'} ) },
\end{equation} 
where $D^*_q$ is the entire collection of positive documents to $q$, including $d^*$ and $H_q$, $D'_q$ comprises $d$ $(d^*$ or $H_q)$ and the negative documents to $q$. 


Knowing that the LLM-based generator can provide precise assessment of relevance due to its inherent re-ranking capability \cite{sun2023chatgpt}, we leverage its feedback for fine-grained training of retriever. Particularly, we apply the following template $\mathcal{T}$: ``\textit{Query: \{q\}. Doc [1]: \{d\_1\}, Doc [2]: \{d\_2\}, \ldots Rank these documents based on their relevance to the query.}'', and obtain the generator's ranking list ``$D_q = d_1, d_2, \dots, d_N$''. Based on this feedback,  we define the following loss function: 
\begin{align}
    \mathcal{L}^{dst} = - \frac{1}{\left | D_q \right |} \sum_{d_k = d_1}^{d_N}  \log \frac{ \exp( \boldsymbol{v}_q^T \boldsymbol{v}_{d_k} )}{\sum\limits_{D'_{q,k}} \exp( \boldsymbol{v}_q^T \boldsymbol{v}_{d'} ) }, 
\end{align}
which follows a variational form of knowledge distillation. Here, $D'_{q,k} = d_{k}, ..., d_N$ indicates $d_k$ and the lower ranked documents of $d_k$. By minimizing the above loss, the retriever is reinforced to discriminate the documents preferred by the generator.



\begin{table*}[ht]
    \centering
    \resizebox{\linewidth}{!}{%
    \begin{tabular}{l|cccccccc|c|ccccc|c}
        \toprule
         & \multicolumn{9}{c|}{BEIR} & \multicolumn{6}{c}{AIR-Bench} \\
        \cmidrule(lr){2-10} \cmidrule(lr){11-16}
         & FiQA & Scidocs & Fever & Arguana & Scifact & T-Covid & Touche & DBPedia & \textbf{AVG} & Law & News & Health & Finance & ArXiv & \textbf{AVG} \\
        \midrule 
        \multicolumn{16}{l}{Contriever} \\
        \midrule
        Contriever & 24.5 & 14.9 & 68.2 & 37.9 & 64.9 & 27.3 & 16.7 & 29.2 & 35.4 & 13.2 & 36.2 & 34.3 & 36.2 & 23.0 & 28.6 \\
        \midrule
        HyDE & 26.2 & 12.4 & 69.5 & 40.9 & 66.5 & 59.7 & 15.8 & 33.9 & 40.6 & 11.0 & 31.4 & 33.9 & 27.7 & 21.8 & 25.2 \\
        Doc2query & 25.5 & 15.3 & 69.4 & 39.5 & 65.9 & 29.2 & 12.6 & 29.9 & 35.9 & 12.7 & 37.8 & 34.3 & 37.4 & 23.1 & 29.1 \\
        QGen & 31.6 & 17.8 & 66.9 & 50.1 & 69.0 & 63.3 & 19.0 & 34.3 & 44.0 & 28.0 & 44.9 & 42.2 & 43.6 & 33.4 & 38.4 \\
        GPL & 33.2 & 17.7 & 77.5 & 44.5 & 58.4 & 67.1 & 21.6 & 42.7 & 45.3 & 24.7 & 44.9 & 44.6 & 41.7 & 36.5 & 38.5 \\
        HyDE+QGen & 34.5 & 17.0 & 70.9 & 49.9 & 68.2 & 69.3 & 21.2 & 38.1 & 46.1 & 21.9 & 40.9 & 31.0 & 36.0 & 29.0 & 31.7 \\
        HyDE+GPL & 35.0 & 17.3 & 76.6 & 42.3 & 57.3 & 74.9 & 25.1 & 42.3 & 46.4 & 18.7 & 40.9 & 40.4 & 32.1 & 32.5 & 32.9 \\
        Reinforced-IR & \textbf{36.8} & \textbf{19.2} & \textbf{81.3} & \textbf{52.6} & \textbf{70.9} & \textbf{78.6} & \textbf{31.1} & \textbf{47.5} & \textbf{52.3} & \textbf{28.4} & \textbf{47.6} & \textbf{45.3} & \textbf{46.1} & \textbf{38.4} & \textbf{41.2} \\
        \midrule
        \multicolumn{16}{l}{BGE-M3} \\
        \midrule
        BGE-M3 & 41.1 & 16.4 & 81.0 & 54.1 & 64.2 & 54.7 & 22.3 & 39.8 & 46.7 & 25.6 & 50.8 & 49.1 & 46.0 & 37.4 & 41.8 \\
        \midrule
        HyDE & 39.2 & 16.9 & 75.8 & 53.2 & 67.2 & 71.7 & 19.6 & 42.5 & 48.3 & 20.7 & 45.6 & 44.1 & 42.7 & 31.6 & 36.9 \\
        Doc2query & 37.7 & 16.8 & 74.8 & 56.0 & 64.3 & 39.4 & 14.0 & 39.5 & 42.8 & 23.8 & 48.4 & 44.7 & 45.7 & 37.8 & 40.1 \\
        QGen & 41.8 & 18.3 & 80.3 & 65.6 & 67.8 & 70.1 & 22.1 & 41.4 & 50.9 & 32.2 & 50.7 & 45.3 & 48.3 & 38.2 & 42.9 \\
        GPL & 43.2 & 18.7 & 79.1 & 65.5 & 65.3 & 74.6 & 24.1 & 41.5 & 51.5 & 27.5 & 50.6 & 51.0 & 45.6 & 37.7 & 42.5 \\
        HyDE+QGen & 42.0 & 19.2 & 75.7 & 60.4 & 68.6 & 78.6 & 22.1 & 42.4 & 51.1 & 27.8 & 46.2 & 40.7 & 46.5 & 31.5 & 38.5 \\
        HyDE+GPL & 40.8 & 19.4 & 75.6 & 58.0 & 67.4 & 78.3 & 23.4 & 42.4 & 50.7 & 20.5 & 44.9 & 44.7 & 41.3 & 30.5 & 36.4 \\
        Reinforced-IR & \textbf{45.8} & \textbf{19.2} & \textbf{84.7} & \textbf{65.1} & \textbf{68.2} & \textbf{83.9} & \textbf{32.4} & \textbf{45.5} & \textbf{55.6} & \textbf{32.5} & \textbf{52.6} & \textbf{51.5} & \textbf{48.9} & \textbf{39.7} & \textbf{45.0} \\
        \bottomrule
    \end{tabular}%
    }
    \vspace{-0.2cm}
    \caption{Overall evaluation (nDCG@10 [\%]) based on BEIR and AIR-Bench datasets.}
    \vspace{-0.3cm} 
    \label{tab:1}
\end{table*}

\section{Experiment}

The experiments are performed for the following research problems. \textbf{RQ 1}. Can Reinforced-IR effectively improve the cross-domain performance over the base retriever? \textbf{RQ 2}. Can Reinforced-IR outperform existing domain-adaptation methods? \textbf{RQ 3}. Whether Reinforced-IR is generally effective with different datasets and model options? \textbf{RQ 4}. Whether the proposed technical designs substantially contribute to the ultimate performance? 

Following the settings in HyDE, We adopt Contriever \cite{izacard2021unsupervised} as our default retriever. Because Contriever is a pre-trained model from unlabeled data, it provides an ideal option to analyze the domain-adaptation effect \cite{gao2022precise}. We also consider Contriever-ft and RetroMAE \cite{xiao2022retromae}, which are fine-tuned from MSMARCO \cite{bajaj2016ms}, as well as BGE M3 \cite{chen2024bge}, GTE \cite{li2023towards}, Stella \cite{zhang2024jasper}, which are fine-tuned from various labeled datasets. 
We leverage Llama-3-8B as our default generator, which is one of the strongest sub 10B LLMs at the time of this paper \cite{dubey2024llama}. We perform extended analysis using both similarly sized LLMs, like Mistral-7B \cite{jiang2023mistral} and Qwen-2.5-7B \cite{hui2024qwen2}, as well as larger and stronger models, including Qwen-2.5-72B, Llama-3-70B, and GPT-4o-mini\footnote{Contriever and Llama-3-8B are set as the default combination of retriever and generator unless specific declaration.}. 


We evaluate the experiment result with two dataset sources. The first one comprises eight low-resource datasets from BEIR \cite{thakur2021beir}. These datasets have not been fine-tuned by any of the retrievers used in the experiments, making them suitable for assessing cross-domain retrieval performance \cite{gao2022precise}. The second source includes five domain-specific datasets from Air-Bench development sets \cite{chen2024air}. As these datasets were recently produced, they have not been used for training either the retrievers or the generators involved in the experiments.

\begin{table*}[ht]
    \centering
    \resizebox{\linewidth}{!}{%
    \begin{tabular}{l|cccccccc|c|ccccc|c}
        \toprule
         & \multicolumn{9}{c|}{BEIR} & \multicolumn{6}{c}{AIR-Bench} \\
        \cmidrule(lr){2-10} \cmidrule(lr){11-16}
         & FiQA & Scidocs & Fever & Argu. & Scifact & T-Covid & Touche & DBPedia & \textbf{AVG} & Law & News & Health & Fin. & ArXiv & \textbf{AVG} \\ 
        \midrule
        Contriever & 24.5 & 14.9 & 68.2 & 37.9 & 64.9 & 27.3 & 16.7 & 29.2 & 35.5 & 13.2 & 36.2 & 34.3 & 36.2 & 23.0 & 28.6 \\
        HyDE (Mist, Ctrv) & 25.3 & 12.4 & 69.5 & 38.6 & 69.2 & 55.0 & 19.7 & 35.8 & 40.7 & 11.3 & 33.4 & 32.5 & 28.6 & 21.6 & 25.5 \\
        HyDE (Qwen, Ctrv) & 24.6 & 13.2 & 63.1 & 39.0 & 68.2 & 52.1 & 17.8 & 31.6 & 38.7 & 11.0 & 32.2 & 33.8 & 29.2 & 22.8 & 25.8 \\
        HyDE (Llama, Ctrv) & 26.2 & 12.4 & 69.5 & 40.9 & 66.5 & 59.7 & 15.8 & 33.9 & 40.6 & 11.0 & 31.4 & 33.9 & 27.7 & 21.8 & 25.2 \\
        Ours (Mist, Ctrv) & \textbf{38.0} & \textbf{19.5} & 81.0 & 52.4 & 70.5 & 78.3 & 30.6 & \textbf{47.5} & 52.2 & 29.7 & \textbf{48.2} & \textbf{46.3} & \textbf{46.8} & 38.3 & \textbf{41.9} \\
        Ours (Qwen, Ctrv) & 36.7 & 19.4 & 78.2 & 52.0 & \textbf{71.1} & 75.7 & \textbf{31.2} & 46.2 & 51.3 & \textbf{30.1} & 46.9 & 45.5 & 46.1 & 37.8 & 41.3 \\
        Ours (Llama, Ctrv) & 36.8 & 19.2 & \textbf{81.3} & \textbf{52.6} & 70.9 & \textbf{78.6} & 31.1 & \textbf{47.5} & \textbf{52.3} & 28.4 & 47.6 & 45.3 & 46.1 & \textbf{38.4} & 41.2 \\
        \midrule
        RetroMAE & 31.6 & 15.0 & 77.3 & 43.4 & 65.3 & 77.2 & 23.7 & 39.0 & 46.6 & 14.5 & 45.7 & 44.7 & 41.0 & 34.5 & 36.1 \\
        HyDE (Llama, Ret) & 26.4 & 13.7 & 75.9 & 37.4 & 62.8 & 72.0 & 25.4 & 38.1 & 44.0 & 11.8 & 44.2 & 29.2 & 37.4 & 26.5 & 29.8 \\
        Ours (Llama, Ret) & \textbf{37.5} & \textbf{17.9} & \textbf{86.2} & \textbf{56.2} & \textbf{70.0} & \textbf{82.6} & \textbf{32.7} & \textbf{46.2} & \textbf{53.7} & \textbf{29.7} & \textbf{52.0} & \textbf{46.1} & \textbf{47.3} & \textbf{38.2} & \textbf{42.6} \\
        \midrule
        Contriever-ft & 32.9 & 16.5 & 75.8 & 44.6 & 67.7 & 59.6 & 20.4 & 41.3 & 44.9 & 13.3 & 46.3 & 45.3 & 43.0 & 32.8 & 36.1 \\
        HyDE (Llama, C-ft) & 31.2 & 15.8 & 77.4 & 40.6 & 67.7 & 72.9 & 26.6 & 41.8 & 46.8 & 12.4 & 45.3 & 39.1 & 40.3 & 29.4 & 33.3 \\
        Ours (Llama, C-ft) & \textbf{38.7} & \textbf{18.7} & \textbf{84.4} & \textbf{52.1} & \textbf{70.4} & \textbf{78.5} & \textbf{34.3} & \textbf{46.4} & \textbf{52.9} & \textbf{30.4} & \textbf{48.3} & \textbf{49.6} & \textbf{48.3} & \textbf{37.4} & \textbf{42.8} \\
        \midrule
        GTE-large & 44.6 & 23.4 & \textbf{84.5} & 57.3 & \textbf{74.3} & 70.2 & 25.5 & 42.4 & 52.8 & 16.1 & 46.0 & 51.5 & 43.0 & 36.7 & 38.9 \\
        HyDE (Llama, gte) & 43.5 & \textbf{23.6} & 81.0 & 53.9 & 75.4 & 75.5 & 22.5 & 44.8 & 52.5 & 13.6 & 44.5 & 47.6 & 40.5 & 34.3 & 36.1 \\
        Ours (Llama, gte) & \textbf{46.1} & 23.2 & 84.1 & \textbf{66.8} & 73.8 & \textbf{84.8} & \textbf{31.7} & \textbf{47.3} & \textbf{57.2} & \textbf{30.3} & \textbf{52.8} & \textbf{56.4} & \textbf{48.3} & \textbf{42.7} & \textbf{46.1} \\
        \midrule
        Stella-base-en-v2 & 38.6 & 18.6 & 79.1 & 60.7 & 72.5 & 64.7 & 21.9 & 39.7 & 49.5 & 15.9 & 42.7 & 50.0 & 40.7 & 30.4 & 36.0 \\
        HyDE (Llama, stella) & 37.8 & 21.2 & 71.5 & 55.1 & 73.6 & 80.4 & 25.3 & 42.2 & 50.9 & 13.3 & 43.3 & 46.4 & 39.2 & 30.8 & 34.6 \\
        Ours (Llama, stella) & \textbf{43.7} & \textbf{22.3} & \textbf{84.2} & \textbf{64.7} & \textbf{74.1} & \textbf{84.2} & \textbf{30.1} & \textbf{45.4} & \textbf{56.1} & \textbf{27.3} & \textbf{49.4} & \textbf{53.2} & \textbf{47.1} & \textbf{37.8} & \textbf{43.0} \\
        \midrule
        BGE-M3 & 41.1 & 16.4 & 81.0 & 54.1 & 64.2 & 54.7 & 22.3 & 39.8 & 46.7 & 25.6 & 50.8 & 49.1 & 46.0 & 37.4 & 41.8 \\
        HyDE (Llama, M3) & 39.2 & 16.9 & 75.8 & 53.2 & 67.2 & 71.7 & 19.6 & 42.5 & 48.3 & 20.7 & 45.6 & 44.1 & 42.7 & 31.6 & 36.9 \\
        Ours (Llama, M3) & \textbf{45.8} & \textbf{19.2} & \textbf{84.7} & \textbf{65.1} & \textbf{68.2} & \textbf{83.9} & \textbf{32.4} & \textbf{45.5} & \textbf{55.6} & \textbf{32.5} & \textbf{52.6} & \textbf{51.5} & \textbf{48.9} & \textbf{39.7} & \textbf{45.0} \\
        \bottomrule
    \end{tabular}%
    }
    \vspace{-0.2cm}
    \caption{Extended evaluation based on additional generators and retrievers.} 
    \vspace{-0.3cm}
    \label{tab:exp-2}
\end{table*}

\begin{table*}[ht]
    \centering
    \resizebox{\linewidth}{!}{%
    \begin{tabular}{l|cccccccc|c|ccccc|c}
        \toprule
         & \multicolumn{9}{c|}{BEIR} & \multicolumn{6}{c}{AIR-Bench} \\
        \cmidrule(lr){2-10} \cmidrule(lr){11-16}
         & FiQA & Scidocs & Fever & Argu. & Scifact & T-Covid & Touche & DBPedia & \textbf{AVG} & Law & News & Health & Fin. & ArXiv & \textbf{AVG} \\
        \midrule
        \multicolumn{16}{l}{Contriever} \\
        \midrule
        \multicolumn{16}{l}{\textit{HyDE}} \\
        \quad Llama3-70B & 28.1 & 14.6 & 74.4 & 40.6 & 69.6 & 51.1 & 19.7 & 36.0 & 41.8 & 12.8 & 36.5 & 35.8 & 32.2 & 24.6 & 28.4 \\
        \quad Qwen2.5-72B & 25.5 & 14.1 & 77.7 & 46.6 & 70.1 & 56.8 & 18.3 & 35.1 & 43.0 & 11.6 & 31.2 & 32.3 & 28.3 & 23.2 & 25.3 \\
        \quad GPT-4o-mini & 26.1 & 13.2 & 76.7 & 44.4 & 68.2 & 57.3 & 18.8 & 33.4 & 42.3 & 12.0 & 34.1 & 34.6 & 31.7 & 24.0 & 27.3 \\
        \multicolumn{16}{l}{\textit{HyDE+QGen}} \\
        \quad Llama3-70B & 35.9 & 18.3 & 75.9 & \textbf{51.1} & \textbf{72.8} & 69.8 & 24.1 & 40.5 & 48.6 & 24.0 & 44.0 & 37.9 & 40.0 & 32.4 & 35.7 \\
        \quad Qwen2.5-72B & 35.7 & 18.0 & 79.3 & 55.1 & 72.4 & 71.0 & 22.6 & 38.4 & 49.1  & 22.4 & 37.7 & 33.7 & 34.8 & 30.2 & 31.8 \\
        \quad GPT-4o-mini & 35.8 & 17.5 & 78.4 & 53.4 & 71.2 & 70.3 & 23.8 & 36.7 & 48.4  & 24.5 & 43.0 & 36.0 & 39.4 & 30.9 & 34.8 \\
        \multicolumn{16}{l}{\textit{HyDE+GPL}} \\
        \quad Llama3-70B & 36.3 & 18.4 & 80.9 & 42.9 & 62.9 & 73.7 & 29.0 & 44.1 & 48.5 & 21.1 & 41.9 & 42.4 & 34.7 & 34.4 & 34.9 \\
        \quad Qwen2.5-72B & 35.9 & 18.5 & \textbf{83.2} & 47.1 & 63.9 & 73.0 & 25.6 & 41.3 & 48.6  & 19.7 & 35.8 & 38.3 & 31.2 & 32.8 & 31.6 \\
        \quad GPT-4o-mini & 35.2 & 17.5 & \textbf{83.2} & 44.9 & 63.3 & 73.7 & 27.6 & 40.9 & 48.3 & 20.9 & 41.1 & 41.6 & 34.7 & 34.2 & 34.5 \\
        \midrule
        Reinforced-IR & \textbf{36.8} & \textbf{19.2} & 81.3 & 52.6 & 70.9 & \textbf{78.6} & \textbf{31.1} & \textbf{47.5} & \textbf{52.3} & \textbf{28.4} & \textbf{47.6} & \textbf{45.3} & \textbf{46.1} & \textbf{38.4} & \textbf{41.2} \\
        \midrule
        \multicolumn{16}{l}{BGE-M3} \\
        \midrule
        \multicolumn{16}{l}{\textit{HyDE}} \\
        \quad Llama3-70B & 40.9 & 17.5 & 80.8 & 54.4 & 70.2 & 68.9 & 22.5 & 43.6 & 49.9 & 22.1 & 46.2 & 44.2 & 40.7 & 32.8 & 37.2 \\
        \quad Qwen2.5-72B & 40.6 & 16.9 & 83.9 & 55.1 & 71.2 & 68.6 & 20.8 & 42.9 & 50.0 & 20.9 & 39.5 & 41.9 & 36.2 & 30.4 & 33.8 \\
        \quad GPT-4o-mini & 40.3 & 16.9 & 83.9 & 52.7 & 69.5 & 72.4 & 21.7 & 42.8 & 50.0 & 22.9 & 44.7 & 44.8 & 41.1 & 31.9 & 37.1 \\
        \multicolumn{16}{l}{\textit{HyDE+QGen}} \\
        \quad Llama3-70B & 43.8 & 19.8 & 79.6 & 62.9 & \textbf{71.9} & 75.6 & 26.4 & 43.5 & 52.9 & 28.0 & 45.8 & 42.2 & 44.4 & 33.3 & 38.7 \\
        \quad Qwen2.5-72B & 42.3 & 20.0 & 82.9 & 60.4 & 71.1 & 74.5 & 25.9 & 43.0 & 52.5 & 21.0  & 39.3 & 42.0  & 36.8 & 31.1 & 34.0 \\
        \quad GPT-4o-mini & 42.7 & 19.5 & 83.1 & 57.5 & 71.1 & 81.0 & 28.4 & 42.4 & 53.2 & 22.7 & 44.6 & 45.2 & 40.0 & 31.8 & 36.9 \\
        \multicolumn{16}{l}{\textit{HyDE+GPL}} \\
        \quad Llama3-70B & 42.8 & \textbf{20.4} & 79.9 & 60.4 & 70.3 & 77.7 & 28.7 & 44.0 & 53.0 & 22.4 & 45.1 & 45.0 & 40.8 & 32.4 & 37.1 \\
        \quad Qwen2.5-72B & 39.2 & 20.0 & 75.6 & 60.3 & 63.8 & 69.0 & 25.4 & 42.5 & 49.5 & 21.3 & 39.7 & 41.6 & 36.0 & 31.2 & 34.0 \\
        \quad GPT-4o-mini & 40.2 & 19.6 & 74.3 & 57.2 & 68.1 & 74.3 & 27.2 & 42.6 & 50.4 & 22.6 & 44.9 & 44.7 & 40.3 & 32.0 & 36.9 \\
        \midrule
        Reinforced-IR & \textbf{45.8} & 19.2 & \textbf{84.7} & \textbf{65.1} & 68.2 & \textbf{83.9} & \textbf{32.4} & \textbf{45.5} & \textbf{55.6} & \textbf{32.5} & \textbf{52.6} & \textbf{51.5} & \textbf{48.9} & \textbf{39.7} & \textbf{45.0} \\ 
        \bottomrule
    \end{tabular}%
    }
    \vspace{-0.2cm}
    \caption{Extended evaluation based on larger LLMs.} 
    \vspace{-0.3cm}
    \label{tab:exp-3}
\end{table*}


\subsection{Experiment Analysis} 
The experiment results are analyzed in comparison with two classes of baselines. The first class relies on generative augmentation, including HyDE \cite{gao2022precise}, which augments the query with hypothetical documents, and Doc2query \cite{nogueira2019document}, which augments the document with pseudo queries. The second class leverages continual fine-tuning, including QGen \cite{ma2020zero,thakur2021beir}, which fine-tunes the retriever based on synthetic queries obtained from the target domain by contrastive learning, and GPL, which performs knowledge distillation for fine-grained training. For the sake of fair comparison, all methods (baselines and Reinforced-IR) are applied to the same set of synthetic queries and the same generator and retriever backbones.

\subsubsection{Overall Evaluation} 
The overall evaluation is demonstrated in Table \ref{tab:1}, where the following analysis is made.  

$\bullet$ \textit{Improvement over base retrievers}. Reinforced-IR substantially improves the base retrievers' cross-domain retrieval performances across all datasets. This effect is particularly evident with Contriever, a pre-trained model from massive unlabeled data. Specifically, it enables the average performance to be improved from 35.4 to 52.3 on BEIR. Moreover, it achieves even larger improvements on AIR-Bench, with the average performance raised from 28.6 to 41.2. Although another base retriever, BGE M3, has been broadly fine-tuned with various question answering datasets, Reinforced-IR still contributes to its performance, increasing its average performance from 46.7 to 55.6 on BEIR, and from 41.8 to 45.0 on AIR-Bench, respectively. The improvement on BGE M3 is remarkable, considering that a broadly fine-tuned retriever has already gained a good command of necessary knowledge on the target domain, where traditional domain adaptation methods struggle to make further improvements \cite{gao2022precise}. 

$\bullet$ \textit{Improvements over domain-adaptation baselines}. Reinforced-IR also demonstrates significant advantages over existing domain-adaptation methods. Notably, it outperforms both generative augmentation methods (HyDE, Doc2Query) and continual fine-tuning methods (QGen, GPL) individually, as well as the combination of the two methods (HyDE+QGen, HyDE+GPL). A closer analysis of the experimental results reveals that the baseline methods struggle to deliver consistent improvements across different datasets and base retrievers. For instance, while HyDE enhances Contriever’s performance on BEIR, it contributes little to BGE M3, as the latter has already undergone extensive fine-tuning. Besides, the benefits of more advanced fine-tuning operations, such as those employed in GPL, are significantly diminished when applied to BGE M3. Additionally, the native combination of HyDE and fine-tuning based methods yields minimal benefit, probably due to the discrepancy between the two strategies. These observations highlight the the limitations of existing domain-adaptation methods, particularly regarding their applicability and significance. In contrast, Reinforced-IR addresses these challenges through its self-boosting mechanism, which effectively mitigates such issues and drives substantial progress.

\begin{table*}[h]
    \centering
    \resizebox{\linewidth}{!}{%
    \begin{tabular}{l|cccccccc|c|ccccc|c}
        \toprule
        & \multicolumn{9}{c|}{BEIR} & \multicolumn{6}{c}{AIR-Bench} \\
        \cmidrule(lr){2-10} \cmidrule(lr){11-16} 
        & FiQA & Scidocs & Fever & Argu. & Scifact & T-Covid & Touche & DBPedia & {AVG} & Law & News & Health & Fin. & ArXiv & {AVG} \\
        \midrule
        Base model & 24.5 & 14.9 & 68.2 & 37.9 & 64.9 & 27.3 & 16.7 & 29.2 & 35.4 & 13.2  & 36.2 & 34.3 & 36.2 & 23.0 & 28.6 \\
        Ret: 0, Gen: 1 & 28.9 & 15.6 & 74.0 & 40.3 & 67.1 & 42.9 & 20.3 & 34.9 & 40.5 & 13.6 & 39.3 & 36.8 & 34.2 & 24.9 & 29.8 \\
        Ret: 1, Gen: 1 & 29.8 & 17.6 & 72.2 & 43.2 & 69.8 & 64.6 & 25.3 & 35.1 & 44.7 & 18.1 & 36.6 & 38.4 & 40.7 & 28.7 & 32.5 \\
        Ret: 1, Gen: 2 & 30.8 & 17.7 & 74.9 & 42.6 & 69.9 & 69.4 & 25.2 & 35.7 & 45.8 & 18.5 & 38.0 & 38.3 & 41.2 & 29.6 & 33.1 \\
        Ret: 2, Gen: 2 & 33.5 & 18.8 & 78.2 & 49.1 & 70.4 & 74.8 & 30.4 & 43.9 & 49.9 & 25.6 & 45.0 & 43.3 & 44.8 & 37.1 & 39.2 \\
        Ret: 2, Gen: 3 & 33.6 & 19.1 & 79.1 & 49.7 & 70.5 & 73.6 & 31.0 & 45.2 & 50.2 & 26.2 & 45.8 & 43.6 & 44.8 & 37.4 & 39.6 \\
        Ret: 3, Gen: 3 & 36.8 & 19.2 & 81.3 & 52.6 & 70.9 & 78.8 & 31.1 & 47.5 & 52.3 & 28.4 & 47.6 & 45.3 & 46.1 & 38.4 & 41.2 \\
        \bottomrule
    \end{tabular}%
    }
    \vspace{-0.2cm}
    \caption{Impact from iterative optimization of generator (Gen) and retriever (Ret).}
    \vspace{-0.3cm}
    \label{tab:exp-4}
\end{table*}

\subsection{Extended Evaluation}
We conduct the extended experiments to explore Reinforced-IR's effectiveness under more situations. 

$\bullet$ \textit{Analysis of different backbones}. We study the impact from using different generators and retrievers, as demonstrated in Table \ref{tab:exp-2}. Our experiment compares three LLMs of similar sizes, including Llama-3 8B (Llama), Qwen-2.5 7B (Qwen), and Mistral 7B (Mist). Besides, we also consider the following types of retrievers: 1) pre-trained model: Contriever, 2) models fine-tuned only with MS MARCO: Contriever-ft, RetroMAE, and 3) broadly fine-tuned models with various datasets: BGE M3, GTE (large), and Stella-v2. From these evaluations, we derive several key observations. 

First, Reinforced-IR consistently outperforms both the base retriever and HyDE baseline when working with different LLM backbones. Despite some underlying differences, e.g., LLama-3 and Mistral are pre-trained more comprehensively than Qwen-2.5, and Qwen-2.5 is more of a bi-lingual LLM compared to the other two models, all methods converge to a superior performance through Reinforced-IR. 
In contrast, none of the HyDE alternatives surpasses Contriever on AIR-Bench, underscoring the incapability of existing generative augmentation methods in dealing with new tasks. 

Second, Reinforced-IR achieves a significant advantage over the baselines when applied to pre-trained and MS MARCO-finetuned retrievers. This result highlights Reinforced-IR's effect in enhancing cross-domain retrieval performance. Moreover, Reinforced-IR also makes substantial contributions to the broadly finetuned retrievers, particularly on AIR-Bench datasets, which demonstrates its generally applicability across diverse application scenarios. 

$\bullet$ \textit{Analysis of larger LLMs}. We incorporate three powerful LLMs to the experiment (Table \ref{tab:exp-4}): Llama3-70B and Qwen2.5-72B, and GPT-4o-mini. This allows us to explore the optimal effect of the existing generative augmentation methods. Our evaluation includes both HyDE and its combinations with other approaches. 

Our experimental results reveal that the use of powerful LLMs can enhance baseline performance in certain scenarios, such as Contriever and BGE M3's retrieval performance on BEIR. However, these improvements are inconsistent across different datasets. Besides, there remains a large performance gap between these methods and Reinforced-IR in most cases. These results highlight that 
it's not enough to simply count on the increased capacity of LLMs. Instead, they underscore the necessity of jointly adapting LLMs and retrievers to optimize the cross-domain retrieval performance. 

$\bullet$ \textit{Analysis of self-boosting}. 
We evaluate the impact of self-boosting by analyzing Reinforced-IR's performance growth throughout the training process (Table \ref{tab:exp-4}). Specifically, the complete set of training queries is divided into three subsets. For each subset, we conduct one self-boosting iteration, consisting of a round of generator optimization via RLRF (Gen-$i$), followed by a round of retriever optimization through RLGF (Ret-$i$). 

The experimental results demonstrate that both self-boosting operations contributes substantially to the improvement of retrieval performance. In each iteration, the optimization of the generator enables the production of more effective query augmentations for the current retriever, improving the performance from ``\textit{Ret}-$i$, \textit{Gen}-$i$'' to ``\textit{Ret}-$i$, \textit{Gen}-($i$+1)''. While the optimization of retriever allows it to make better use of the augmented queries from the current generator, which further improves the performance from ``\textit{Ret}-$i$, \textit{Gen}-($i$+1)'' to ``\textit{Ret}-($i$+1), \textit{Gen}-($i$+1)''. This iterative refinement ultimately results in Reinforced-IR's superior performance across the entire training dataset. 

\subsubsection{Ablation Study} 
We make detailed analysis for Reinforced-IR's technical factors with the ablation study in Table \ref{tab:ablation}. 

$\bullet$ \textit{Training methods}. In our experiment, we replace the original DPO with supervised fine-tuning for the generator's training, using the winning candidate as the supervision label (w/o DPO). Additionally, we substituted basic contrastive learning for knowledge distillation during retriever's training (w/o Distillation). The experiment result shows that both modifications lead to significant decline of empirical performance on the two evaluation benchmarks. This decline is attributed to the alternative methods' inability to incorporate the fine-grained feedback from the generator and retriever, specifically, the usability of augmented queries and document relevance, thereby hindering the effective utilization of training data.  

$\bullet$ \textit{Impact of proximity objective}. We further replace the proximity objective in Eq. \ref{eq:6} with the basic objective: $\alpha_0 * \boldsymbol{v}_q^T \boldsymbol{v}_{d} + \sum_{L} \alpha_i * \boldsymbol{v}_{h_i}^T \boldsymbol{v}_{d}$. As discussed, this alternative form requires the model to accomplish both query-to-doc and doc-to-doc matching, thus increasing the training difficulty. Our experiment result verifies proximity objective's overall effectiveness in general scenarios, as the alternative method (w/o Proximity) significantly reduces the performance on AIR-Bench, which solely comprises question-answering style tasks, while leading to a neural impact on BEIR, which constitutes miscellaneous tasks. 

$\bullet$ \textit{Candidate filtering}. We disable the filtering rules in Eq. \ref{eq:4} and make direct use of the unfiltered candidates (w/o Filtering rule-1, w/o Filtering rule-2). The experiment result highlights the significance of both rules on BEIR's performance. This can be attributed to BEIR's diverse retrieval tasks, which increases the likelihood of generating unsuitable query augmentation from the generator. As such, the filtering operations are essential to optimizing the performance. In contrast, AIR-Bench focuses solely on question-answering tasks, allowing for more reliable query augmentation and diminishing the need for candidate filtering.  


\begin{table}[t]
    \centering
    \resizebox{\linewidth}{!}{
    \begin{tabular}{l|cc|cc}
        \toprule
         & BEIR & $\Delta$ & AIR-Bench & $\Delta$ \\
        \midrule
        Reinforced-IR (default) & 52.3 & - & 41.2 & - \\
        \midrule
        \quad w/o DPO & 50.6 & -1.7 & 40.6 & -0.6 \\
        \quad w/o Distillation & 49.3 & -3.0 & 38.9 & -2.3 \\         
        \quad w/o Proximity & 52.3 & -0.0 & 40.2 & -1.0 \\
        \quad w/o Filtering rule-1 & 49.0 & -3.3 & 41.1 & -0.1 \\
        \quad w/o Filtering rule-2 & 49.7 & -2.6 & 40.9 & -0.3 \\
        \bottomrule
    \end{tabular}%
    }
    \vspace{-0.2cm}
    \caption{Ablation studies.} 
    \vspace{-0.6cm} 
    \label{tab:ablation}
\end{table}

\section{Related Work} 
Cross-domain retrieval is an important but challenging problem for existing techniques. As demonstrated by the popular benchmarks in this field \cite{thakur2021beir}, the pre-trained retrievers are prone to inferior performances when they are applied directly for a new working scenario. To tackle this challenge, one common strategy is to perform multi-task training, where a pre-trained retriever is broadly fine-tuned using extensive labeled datasets. By learning from diverse tasks during training, the retriever develops a stronger ability to handle new tasks during testing \cite{wang2022text,xiao2024c,su2022one}. However, multi-task retrievers often make trade-offs between individual tasks to optimize overall performance, leading to significant performance gaps when compared to specialized retrievers in target domains. 

Another line of research focuses on the continual fine-tuning of pre-trained retrievers using synthetic data generated from a target domain \cite{ma2020zero,thakur2021beir,wang2021gpl}. These approaches leverage generators to produce synthetic queries for unlabeled documents, which creates training samples to fine-tune the pre-trained models. Thanks to the popularity of language models, it's made possible to produce synthetic queries at scale, \cite{bonifacio2022inpars,jeronymo2023inpars,wang2023improving}, enabling the corresponding approaches to be easily conducted in practice. However, these approaches could deliver limited performance gains due to the potential mismatch between the synthetic data and actual scenarios. 

Different from the fine-tuning methods, generation augmented retrieval (GAR) makes direct use of generation models to address the cross-domain problems \cite{mao2020generation}. These methods enrich query and document with extra information, enabling relevant data to identified in a straightforward way. 
Nowadays, large language models are widely adopted as the backbone generator \cite{gao2022precise,wang2023query2doc}, which contributes to the performance and applicability of corresponding methods. Although GAR is widely perceived as a promising strategy, it's not enough solely rely on general LLMs, as they still lack necessary knowledge required to generate effective query augmentations for highly specialized problems.  

\section{Conclusion} 
\vspace{-0.2cm}
In this paper, we introduce Reinforced-IR, a novel self-boosting framework for cross-domain retrieval. Our method employs two advanced learning algorithms: RLRF and RLGF. These algorithms enable the generator and retriever to mutually reinforce each other through feedback, leading to a progressive enhancement of retrieval performance. The effectiveness of Reinforced-IR is thoroughly validated, as it outperforms existing domain adaptation methods by a huge advantage, delivering superior performance across various application scenarios. 

\bibliography{custom}
\bibliographystyle{acl_natbib}

\clearpage
\newpage
\appendix

\section{Implementation Details}
When get query embeddings, we set $\alpha$ to 0.8. In RLRF, $\gamma$ is sequentially set to 1.05, 1.08, and 1.1 over three iterations. Meanwhile, in RLGF, both $\alpha_0$ and $\alpha_i$ are defined as 0.5.

For LLM, we use the following version:

1) Meta-Llama-3-8B-Instruct, 

2) Qwen-2.5-7B-Instruct, 

3) Mistral-7B-Instruct-v0.3, 

4) Meta-Llama-3-70B-Instruct, 

5) Qwen-2.5-72B-Instruct, 

6) gpt-4o-mini-2024-07-18

\section{Query Generation}
To enhance the effectiveness of query generation, we have implemented refinements to QGen, which encompasses two distinct phases: query generation and quality control.
During the query generation phase, we provide the LLM with relevant documents and tasks, prompting it to produce corresponding queries. In the quality control phase, the tasks, documents, and generated queries are presented to the LLM again, enabling it to evaluate whether the generated queries are relevant to the documents. Queries deemed irrelevant are then filtered out.

Taking Trec-covid dataset as an example, for query generation, we use the following prompt:
\begin{mdframed}[backgroundcolor=gray!20, linecolor=gray]
\small
Here is a retrieval task (Task) and a document (Passage):\\
\\
Task: Given a query on COVID-19, retrieve documents that answer the query.\\
\\
Passage: \{passage\}\\
\\
Given the retrieval task and the document, your mission is:\\
- Generate a query on COVID-19 that the document can answer.\\
\\
Note:\\
- The generated query should not contain the pronouns such as "this", "that", "it", "there", "here", etc.\\
- The generated query should be clear and 5 to 10 words.\\
- The generated query should be common and formal in terms of language style.\\
\\
Your output should be a string of the generated query. Remember do not explain your output.\\
\\
Your output:
\end{mdframed}

For quality control, we use the following prompt:
\begin{mdframed}[backgroundcolor=gray!20, linecolor=gray]
\small
Given a retrieval task (Task), a query (Query), and a document (Passage), your mission is Judge whether the document can answer the query..\\
\\
Task: Given a query on COVID-19, retrieve documents that answer the query.\\
\\
Query: \{query\}\\
Passage: \{passage\}\\
\\
Your output must be one of the following:\\
- 0: No, the document cannot answer the query.\\
- 1: Yes, the document can answer the query.\\
\\
Do not explain your answer in the output. Your output must be a single number.\\
\\
Your output:
\end{mdframed}

\section{Hypothetical Document}
For hypothetical document, we use the following prompt:

\begin{mdframed}[backgroundcolor=gray!20, linecolor=gray]
\small
Given a retrieval task and a query, your mission is to generate a brief document for the query in the context of the retrieval task.\\
\\
Task: Given a query on COVID-19, retrieve documents that answer the query.\\
\\
Query: \{query\}\\
\\
Your output:
\end{mdframed} 

\end{document}